\RequirePackage{fix-cm}
\documentclass[smallcondensed]{svjour3}     
\usepackage{float}
\usepackage{amsmath}
\usepackage{algorithm}
\usepackage{algpseudocode}
\usepackage{amsfonts}
\usepackage [latin1]{inputenc}

\smartqed  
\usepackage{graphicx}
\begin{document}

\title{Gaussian Process-Based Model Predictive
Control for Overtaking in Autonomous Driving
}
\subtitle{Our paper falls on categories (1), (3) and (5).}


\author{Wenjun Liu $^{1}$        \and
        Chang Liu $^{2,1}$         \and
        Guang Chen $^{2,1,*}$        \and
        Peng Hang  $^{3}$        \and
        Alois Knoll$^{1}$
}


\institute{1. Department of Informatics, Technical University of Munich, Munich, Germany \\
           2. College of Automotive Engineering, Tongji University, Shanghai, China \\
           3. School of Mechanical and Aerospace Engineering, Nanyang Technological University, Singapore\\
           Corresponding Author: Guang Chen. \email{guangchen@tongji.edu.cn}
}

\date{Received: date / Accepted: date}

\maketitle

\begin{abstract}
This paper proposes a novel framework for addressing the challenge of autonomous overtaking and obstacle avoidance, which incorporates the overtaking path planning into Gaussian Process-based model predictive control (GPMPC). Compared with the conventional control strategies, this approach has two main advantages. Firstly, combining Gaussian Process (GP) regression with a nominal model allows for learning from model mismatch and unmodeled dynamics, which enhances a simple model and delivers significantly better results. Due to the approximation for propagating uncertainties, we can furthermore satisfy the constraints and thereby safety of the vehicle is ensured. Secondly, we convert the geometric relationship between the ego vehicle and other obstacle vehicles into the constraints. Without relying on a higher-level path planner, this approach substantially reduces the computational burden. In addition, we transform the state constraints under the model predictive control (MPC) framework into a soft constraint and incorporate it as relaxed barrier function into the cost function, which makes the optimizer more efficient. Simulation results indicate that the proposed method can not only fulfill the overtaking tasks but also maintain safety at all times.
\end{abstract}

\section{Introduction}
\label{intro}
Autonomous driving has attracted considerable attention because of its promising future \cite{chehri2019autonomous}. A number of modern techniques have been employed for advanced driving assistant system, such as adaptive cruise control \cite{wu2019cooperative}, automatic parking \cite{ye2019linear}, etc, which can be regarded as the low level autonomous driving. However, due to its demands of high reliability and real-time practicality of fully autonomous driving, performing overtaking maneuvers imposes a major challenge \cite{cha2018op}. Even for human beings, overtaking is also a dangerous task, therefore, reliable and safe autonomous overtaking systems are becoming more and more appealing  \cite{lattarulo2018linear}.

From a control perspective, the essential building blocks for autonomous overtaking can be categorized into trajectory tracking and path planning. The research related to the first group focuses on obtaining accurate model descriptions to achieve high-performance control \cite{hewing2019cautious}. While vehicle dynamics are disreputable difficult to model in complex situations, learning-based control method has been proposed and widely applied to solve this problem. This method digs information from the measurement data and combines with model predictive control (MPC), which also takes constraints into account \cite{hewing2020learning}.  \cite{aswani2013provably} designed a learning-based MPC structure based on this conception, by making use of a nominal model and an additive learned term. The satisfaction of constraints in closed loop is guaranteed due to the robust MPC properties. However, this method requires a hard constraint to uncertainty, which is conservative in practice. Stochastic MPC with parametric models addresses this issue by considering the probabilistic distribution of uncertainty \cite{soloperto2018learning,koller2018learning}, but these methods suffer from large or unbounded uncertainty, which may lead to unstable in control progress. Another learning-based methods make adaption on controller design. \cite{rosolia2019learning} applied its method to the problem of autonomous racing, where the problem was formed as an iterative control task and the controller was updated in each time step by using input sequence. Gaussian process (GP) is a non-parametric machine learning approach and has shown success in combining with predictive control. For instance, \cite{hewing2018cautious} used GP to estimate the residual model uncertainty for autonomous miniature race cars. They presented an approximation technique for propagating state uncertainties that formulated the chance constrained MPC, which enabled safety through the whole racing.

In order to avoid collisions, there is one view states that the tracking control strategy needs a higher-level path planner. This method was used in \cite{gao2010predictive} but the vehicle modeled as a simple point mass model which neglects the vehicle kinematics and dynamics, this will strictly limit the performance when the speed of vehicle increases. A more complex dynamical model is adopted in the planner in \cite{frazzoli2005maneuver} to generate the reference trajectories for the low level tracking controller. However, it is too complex to be solved due to the mixed-integer program optimization problem. So, this approach is not well suited for real-time overtaking task. Instead, one-level approach have been investigated recently. \cite{liniger2015optimization} combined the path planning and path tracking into one nonlinear optimization problem, the path planner was based on dynamic programming and merged into model predictive contour control. However, they only take the stationary obstacles rather than moving vehicles into account. The complexity of obstacle avoidance would increase when the obstacles is moving, this approach may not work in this situation.  A short-term path planning in \cite{franco2019short} considered both static obstacles and moving vehicles, proposing a flexible overtaking paradigm based on adaptive MPC. Since the bicycle model is only concerned with kinematics, the lateral control with regard to tire model is simplified, the generated trajectories were a bit infeasible.

In this paper, we investigate the autonomous overtaking problems with GP-based MPC approach. A single track model considering the nonlinear wheel dynamics is adopted as the nominal model. GP is used to learn the unknown deviation between the nominal model and the true plant dynamics. Then we convert the geometric relationship between the ego vehicle and other vehicles/obstacles into constraints to achieve overtaking.
Finally, a framework for addressing autonomous overtaking based on GPMPC controller is proposed and compared with the nominal NMPC controller.

The rest of this paper is constructed as follows. The vehicle model is introduced in section 2. In section 3, Gaussian process regression is introduced. In this section, we first present the preliminaries of GP, then give the approach of how to obtain training data, and give an approximate approach for the propagation of uncertainty in multi-step-ahead prediction. In section
4, we design GPMPC controller for vehicle overtaking problem. In section 5, simulations are conducted to verify the effectiveness of the proposed controller. Finally, we
conclude in section 6.

\section{Vehicle Model}
\label{sec:1}
Establishing a reasonable vehicle model is not only a prerequisite for designing a model predictive controller, but also a basis for realizing vehicle overtaking control. Therefore, it is necessary to select control variables according to the driving conditions of the vehicle to establish a kinematics and dynamics model that can accurately describe the vehicle. However, if the model is too complex, it will affect the real-time performance of the control algorithm.
\begin{figure}[ht]
    \centering
    \includegraphics[scale=0.39]{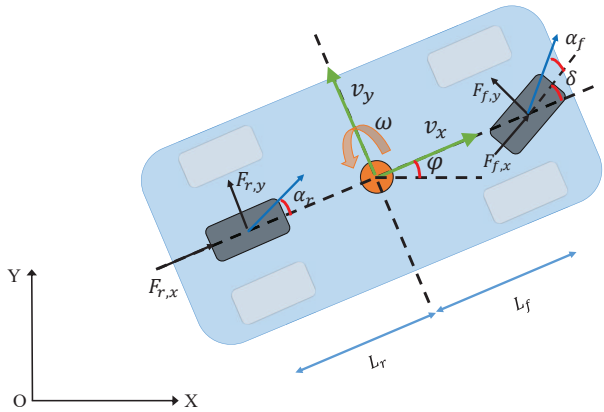}
     \caption{A schematic drawing of the bicycle model}\label{f2.2}
\end{figure}

In this chapter, a simplified vehicle model is introduced to trade off computational performance and vehicle characteristics. A single track model is adopted in this paper as shown in Fig. 1, where each side wheels are merged into one wheel. For better maneuverability and high-speed stability, we assume that only the front wheel can steer. Only the longitudinal and lateral as well as yaw motion will be considered, the pitch and roll dynamics are neglected \cite{langaaker2018cautious}. The vehicle is typically assumed to be a mass point with the global position coordinates $(X,Y)$ and the yaw angle $\varphi$, while $v_x$ and $v_y$ represent the longitudinal and lateral velocities respectively. $\omega$ refers to the yaw rate. The other parameters are vehicle mass $M$, yaw moment of inertia $I_z$, the steering angle $\delta$, the distance between the center of gravity (c.g.) of the vehicle and the front and rear wheel are $L_f$ and $L_r$, respectively. The forces which act on the front and rear wheel in longitudinal and lateral direction are defined by $F_{f,x}$, $F_{f,y}$, $F_{r,x}$, $F_{r,y}$. Finally, the front- and rear-slip angle are $\alpha_f$ and $\alpha_r$, respectively. Then the vehicle model is given by
\begin{equation}
\boldsymbol{f}(\boldsymbol{x}, \boldsymbol{u})=\left[\begin{array}{c}
v_{x} \cos (\varphi)-v_{y} \sin (\varphi) \\
v_{x} \sin (\varphi)+v_{y} \cos (\varphi) \\
\omega \\
\frac{1}{M}\left(F_{r, x}+F_{f, x} \cos (\delta)-F_{f, y} \sin (\delta)+M \omega v_{y}\right) \\
\frac{1}{M}\left(F_{r, y}+F_{f, x} \sin (\delta)+F_{f, y} \cos (\delta)-M \omega v_{x}\right) \\
\frac{1}{I_{z}}\left(F_{f, y} L_{f} \cos (\delta)+F_{f, x} L_{f} \sin (\delta)-F_{r, y} L_{r}\right)
\end{array}\right]
\end{equation}
where $\boldsymbol{x} = [X;Y;\varphi;v_x;v_y;\omega]$ is the state of the system, while the input vector $\boldsymbol{u} = [\delta;T]$ to the system are the steering angle $\delta$ and the acceleration/brake pedal $T$ ($T \in [-1,1]$). The vehicle's velocity is controlled increasing or decreasing by the throttle, when it is set to $T > 0$ or $T < 0$.

The longitudinal wheel forces $F_{f/r,x}$ in vehicle coordinates are modeled simply as proportional to the acceleration/brake pedal $T$ and the torque distribution $\zeta$ by
\begin{equation}
\begin{aligned}
F_{W} &=T\left((T>0) F_a+(T<0) F_b\textrm{sign}\left(V_{x}\right)\right) \\
F_{f, x} &=(1-\zeta) F_{W}\\
F_{r, x} &=\zeta F_{W}
\end{aligned}
\end{equation}
where $F_a$ and $F_b$ are acceleration force and brake force respectively.

According to \cite{pacejka1992magic}, the lateral forces $F_{f,y}$ and $F_{r,y}$ are given by the full MAGIC formulas.
\begin{equation}
    \begin{array}{l}
    F_{f, y}=D_{f} \textrm{sin} \left[C_{f} \textrm{arctan} \left(B_{f} \alpha_{f}-E_{f}\left(B_{f} \alpha_{f}-\textrm{arctan} \left(B_{f} \alpha_{f}\right)\right)\right)\right]\\
F_{r, y}=D_{r} \textrm{sin} \left[C_{r} \textrm{arctan} \left(B_{r} \alpha_{r}-E_{r}\left(B_{r} \alpha_{r}-\textrm{arctan} \left(B_{r} \alpha_{r}\right)\right)\right)\right]
\end{array}
\end{equation}
where $B_*$ is stiffness factor, $D_*$ is peak factor, $C_*$ and $E_*$ are shape factors, $\alpha*$ represents the front wheel slip angle and rear wheel slip angle, respectively.

However, the full MAGIC formulas are too complicated for the practice in some cases. In order to ease the computational burden, the following simplified Pacejka Tire Model \cite{elbanhawi2018receding} is used, which is a linear approximation of (3).
\begin{equation}\label{simpletire}
   \begin{array}{l}
    F_{f, y}=C_{l, f} \alpha_{f}\\
    F_{r, y}=C_{l, r} \alpha_{r}
    \end{array}
\end{equation}
where $C_{l, f}$ and $C_{l, r}$ are the front and rear cornering stiffness. For both equations, the wheel slip angle $\alpha_{*}$ is defined as the angle between the orientation of the tire and the orientation of the velocity vector of the wheel \cite{rajamani2011vehicle}

\begin{equation}\label{slipangle}
\begin{array}{l}
\alpha_{f}=\textrm{arctan}\left(\frac{v_y+L_{f} \dot{\varphi}}{v_x}\right)-\delta \\
\alpha_{r}=\textrm{arctan}\left(\frac{v_y-L_{r} \dot{\varphi}}{v_x}\right)
\end{array}
\end{equation}

In this paper, the model with full MAGIC formulas serves as the true vehicle model, the model with simplified Pacejka Tire Model serves as the nominal model.

\section{Gaussian Process Regression}

\subsection{Preliminaries of Gaussian Process Regression}
As defined in \cite{rasmussen2003gaussian}, a Gaussian process is a collection of random variables,
any finite number of which has a joint Gaussian distribution. For easy identification, the notation of the training data set of the GP is defined as
$$
\begin{aligned}
\mathcal{D}=\{\mathbf{Z}&=\left[\boldsymbol{z}_{1}, \ldots, \boldsymbol{z}_{N}\right] \in \mathbb{R}^{n_{z} \times N} \\
\mathbf{Y} &\left.=\left[y_{1}, \ldots, y_{N}\right] \in \mathbb{R}^{1 \times N}\right\}
\end{aligned}
$$
where $n_{z}$ stands for the dimension of the input vector $\boldsymbol{z}$, $N$ is the number of the input and output pairs $(\boldsymbol{z}_{k},y_{k})$. With input vector $\boldsymbol{z}_{k}$, each output $y_k$ can be represented by $y_{k}=\boldsymbol{d}\left(\boldsymbol{z}_{k}\right)+\varepsilon_{k}$, where $\boldsymbol{d}:\mathbb{R}^{n_{z}} \rightarrow \mathbb{R}$ and $\varepsilon_{k} \sim \mathcal{N}\left(0, \sigma_{\varepsilon}^{2}\right)$ denotes Gaussian measurement noise.
Just like a Gaussian distribution is specified by its mean and variance, a Gaussian process is completely defined by mean function $\boldsymbol{m}(\boldsymbol{z})$ and a covariance function
$\boldsymbol{k}(\boldsymbol{z},\boldsymbol{z}^{\prime})$.
\begin{equation}\label{GPfunction}
 \begin{aligned}
\boldsymbol{m}(\boldsymbol{z}) &=\mathbb{E}[\boldsymbol{d}(\boldsymbol{z})] \\
\boldsymbol{k}\left(\boldsymbol{z}, \boldsymbol{z}^{\prime}\right) &=\mathbb{E}\left[(\boldsymbol{d}(\boldsymbol{z})-m(\boldsymbol{z}))\left(\boldsymbol{d}\left(\boldsymbol{z}^{\prime}\right)-m\left(\boldsymbol{z}^{\prime}\right)\right)\right]
\end{aligned}
\end{equation}
Thus Gaussian process is written as
\begin{equation}\label{gp}
    \boldsymbol{d}(\boldsymbol{z}) \sim \mathcal{G} \mathcal{P}\left(\boldsymbol{m}(\boldsymbol{z}), \boldsymbol{k}\left(\boldsymbol{z}, \boldsymbol{z}^{\prime}\right)\right)
\end{equation}
The covariance function $\boldsymbol{k}(\boldsymbol{z},\boldsymbol{z}^{\prime})$ is also known as kernel function. A squared exponential kernel is adopted in this paper.
\begin{equation}\label{kernel}
    \boldsymbol{k}(\boldsymbol{z},\boldsymbol{z}^{\prime})=\sigma_f^2exp(-\frac{1}{2}(\boldsymbol{z}-\boldsymbol{z}^{\prime})^\textrm{T}\textbf{M}^{-1}(\boldsymbol{z}-\boldsymbol{z}^{\prime}))
\end{equation}
where $\sigma_f^2$ and $\textbf{M}$ are the signal variance and the diagonal matrix of squared characteristic length-scales, respectively. $\textbf{M}=\textrm{diag}\left(\left[\ell_{1}, \ldots,  \ell_{n_{z}}\right]\right)$. Moreover, the noise variance $\sigma_n^2$ is usually to be considered, which can be added directly behind (8). These three parameters are called hyper-parameters, which are collected by parameter vector $\boldsymbol{\theta}=[\ell_1,...,\ell_{n_{z}},\sigma_f^2,\sigma_n^2]$.  With predefined kernel function, we can get the prior distribution of samples. Hyper-parameters have a great influence on the performance of GP. In this paper, the Maximum Likelihood approach is adopted to obtain the optimal hyper-parameters \cite{rasmussen2003gaussian}.

The posterior distribution at the test point $\boldsymbol{z}_{*}$ is also a Gaussian distribution with mean and variance \cite{williams2006gaussian}.
\begin{equation}\label{predictivemean}
\mu^{d}(\boldsymbol{z}_{*})=\textbf{K}_{*}^{\top}\left[\textbf{K}+\sigma_{n}^{2} \textbf{I}\right]^{-1} \textbf{Y}
\end{equation}

\begin{equation}\label{predictivevar}
    \Sigma^{d}(\boldsymbol{z}_{*})=\textbf{K}_{*, *}-\textbf{K}_{*}^{\top}\left[\textbf{K}+\sigma_{n}^{2} \textbf{I}\right]^{-1} \textbf{K}_{*}
\end{equation}
where $d$ denotes the $d$-th dimension of the output. $\textbf{K}$, $\textbf{K}_*$ and $\textbf{K}_{*,*}$ are short for $\textbf{K}(\textbf{Z}, \textbf{Z})$, $\textbf{K}\left(\textbf{Z}, \boldsymbol{z}_{*}\right)$ and $\textbf{K}\left(\boldsymbol{z}_{*}, \boldsymbol{z}_{*}\right)$, respectively. And we have

$[\textbf{K}(\textbf{Z}, \textbf{Z})]_{i j}=\boldsymbol{k}\left(\boldsymbol{z}_{i}, \boldsymbol{z}_{j}\right)$,
$[\textbf{K}(\textbf{Z}, \boldsymbol{z}_{*})]_{ j}=\boldsymbol{k}\left(\boldsymbol{z}_{j}, \boldsymbol{z}_{*}\right)$
and $\textbf{K}(\boldsymbol{z}_{*}, \boldsymbol{z}_{*})=\boldsymbol{k}\left(\boldsymbol{z}_{*}, \boldsymbol{z}_{*}\right)$.

As discussed above, a GP regression for one dimensional output has been presented. In our paper, the output vector has $n_d$ dimensions. The multivariate GP approximation is given by
\begin{equation}\label{predictivevar}
    \boldsymbol{d}(\boldsymbol{z}_{*}) \sim \mathcal{N}\left(\boldsymbol{\mu}^{d}(\boldsymbol{z}_{*}), \boldsymbol{\Sigma}^{d}(\boldsymbol{z}_{*})\right)
\end{equation}
where
$$\boldsymbol{\mu}^{d}(\boldsymbol{z}_{*})=\left[\mu^{1}(\boldsymbol{z}_{*}) ; \ldots ; \mu^{n_{d}}(\boldsymbol{z}_{*})\right]$$
$$\boldsymbol{\Sigma}^{d}({\boldsymbol{z}_{*}})=diag(\left[\Sigma^{1}(\boldsymbol{z}_{*}) ; \ldots ; \Sigma^{n_{d}}(\boldsymbol{z}_{*})\right])$$
\subsection{Training data acquisition for GP}

The true vehicle model is presented as follows:
\begin{equation}\label{learningmodel}
    \boldsymbol{x}_{k+1}=\boldsymbol{f}_{n}\left(\boldsymbol{x}_{k}, \boldsymbol{u}_{k}\right)+\textbf{B}_{d} (\boldsymbol{d}\left(\boldsymbol{x}_{k}, \boldsymbol{u}_{k}\right)+\boldsymbol{w}_k)
\end{equation}%
where $\boldsymbol{f}_{n}\left(\boldsymbol{x}_{k}, \boldsymbol{u}_{k}\right)$ is a nominal function, which is the discrete model of (1). $\boldsymbol{x}_{k} \in \mathbb{R}^{n}$ denotes the state variables and $\boldsymbol{u}_{k} \in \mathbb{R}^{m}$ is the control inputs. The matrix $\textbf{B}_d$ picks the states of system which are affected by the model error. $\boldsymbol{d}$ is the GP to capture the model mismatch and unmodeled dynamics. In our paper, we assume that the model mismatch and unmodeled dynamics, as well as the process noise $\boldsymbol{w}_k$ only affect the longitudinal velocity $v_x$, the lateral velocity $v_y$ and the yaw rate $w$, i.e. $\textbf{B}_d=\left[0 ; \textbf{I}_{3}\right]$. $\boldsymbol{w}_k$ is i.i.d normally distributed process noise with $\boldsymbol{w}_k \sim \mathcal{N}\left(0, \boldsymbol{\Sigma}^{w}\right)$, $\boldsymbol{\Sigma}^{w}=\operatorname{diag}\left[\sigma_{v_{x}}^{2} , \sigma_{v_{y}}^{2} , \sigma_{\omega}^{2}\right]$.

Since Gaussian process is a non-parametric method, measurement data of states and inputs should be collected to infer the GP model. For a specific input data point $\boldsymbol{z}_{k}=\left[\boldsymbol{x}_{k};\boldsymbol{u}_{k}\right]$, we have the training output as follows:
\begin{equation}
{y}_{k}=\boldsymbol{d}\left(\boldsymbol{x}_{k}, \boldsymbol{u}_{k}\right)+\boldsymbol{w}_k=\textbf{B}_{d}^{\dagger}\left(\boldsymbol{x}_{k+1}-\boldsymbol{f}_{n}\left(\boldsymbol{x}_{k}, \boldsymbol{u}_{k}\right)\right)
\end{equation}
where $\textbf{B}_{d}^{\dagger}$ is the Moore-Penrose pseudo-inverse.

Then the training input and output pairs $(\boldsymbol{z}_{k},y_{k})$ will be used to train GP. The performance of GP relies on the training data which adds to the system. The more data we add to the model, the more precise result we can obtain. However, the increase of the size of training data will be a heavy burden to the solver, resulting in computational infeasibility over time. To avoid this situation and keep the training data size at an acceptable level, we restrict the number of actively used data points to $N_{max}$. Once the training data size reaches maximum size $N_{max}$, some data need to be replaced. The data selection method is based on a distance measure $\Theta_{*}$ which has been introduced in \cite{kabzan2019learning}. It is defined as the posterior variance at the data point location $\boldsymbol{z}_{*}$, given all other data points currently in the dictionary $\textbf{Z}_{\backslash *}$, which is shown as:
\begin{equation}
    \Theta_{*}= \textbf{K}_{\boldsymbol{z}_{*},\boldsymbol{z}_{*}}-\textbf{K}_{\boldsymbol{z}_{*},\textbf{Z}_{\backslash *}}(\textbf{K}_{\textbf{Z}_{\backslash *}, \textbf{Z}_{\backslash *}}+\sigma \textbf{I})^{-1}\textbf{K}_{\textbf{Z}_{\backslash *},\boldsymbol{z}_{*}}
\end{equation}
where $\sigma$ is a tuning parameter. The data with the lowest distance measure should be dropped.
\subsection{Approximate Uncertainty Propagation}

The states and GP disturbances are approximated as jointly Gaussian distribution at each time step.
\begin{equation}
\begin{aligned}
& \left[\begin{array}{ll}

\boldsymbol{x}_{k}^\textrm{T} & (\boldsymbol{d}_{k}+\boldsymbol{w}_k)^\textrm{T}
\end{array}\right]^\textrm{T} \sim \mathcal{N}\left(\boldsymbol{\mu}_{k}, \boldsymbol{\Sigma}_{k}\right)\\
&=\mathcal{N}\left(\left[\begin{array}{c}
\boldsymbol{\mu}_{k}^{x} \\
\boldsymbol{\mu}_{k}^{d}
\end{array}\right],\left[\begin{array}{cc}
\boldsymbol{\Sigma}_{k}^{x} & (\boldsymbol{\Sigma}_{k}^{d x})^\textrm{T} \\
\boldsymbol{\Sigma}_{k}^{d x} & \boldsymbol{\Sigma}_{k}^{d}+\boldsymbol{\Sigma}^w

\end{array}\right]\right)
\end{aligned}
\end{equation}
where $\boldsymbol{d}_k$ represents model mismatch and unmodeled dynamics learned by GP, $\boldsymbol{\Sigma}_{k}^{d x}$ denotes the covariances between states and GP. To approximate the distribution of predicted states over the prediction horizon, linearization techniques related to extended Kalman filter are then derived, which allows simple update for the state mean and variance.
\begin{equation}
\boldsymbol{\mu}_{k+1}^{x}=\boldsymbol{f}_{n}\left(\boldsymbol{\mu}_{k}^{x}, \boldsymbol{u}_{k}\right)+\textbf{B}_{d} \boldsymbol{\mu}_{k}^{d} \\
\end{equation}
\begin{equation}\label{eq413}
\boldsymbol{\Sigma}_{k+1}^{x}=\left[\nabla_{x} \boldsymbol{f}_{n}\left(\boldsymbol{\mu}_{k}^{x}, \boldsymbol{u}_{k}\right) \quad \textbf{B}_{d}\right] \boldsymbol{\Sigma}_{k}
\left[\nabla_{x} \boldsymbol{f}_{n}\left(\boldsymbol{\mu}_{k}^{x}, \boldsymbol{u}_{k}\right) \quad \textbf{B}_{d}\right]^\textrm{T}
\end{equation}

In real scenario, the input will not always be deterministic, e.g. in the context of multi-step-ahead prediction, the last time's prediction is the input for the next iteration, which has a probability distribution. The challenge is how to propagate the resulting stochastic state distributions over the prediction horizon \cite{hewing2020learning}. Assume the input itself is Gaussian, $\boldsymbol{z}_{k} \sim \mathcal{N}(\boldsymbol{\mu}^{z}_{k}, \boldsymbol{\Sigma}^{z}_{k})$, the predictive distribution is stated as
\begin{equation}
    p\left(\boldsymbol{d}\left(\boldsymbol{z}_{k}\right) \mid \boldsymbol{\mu}^{z}_{k}, \boldsymbol{\Sigma}^{z}_{k}\right)=\int p\left(\boldsymbol{d}\left(\boldsymbol{z}_{k}\right) \mid \boldsymbol{z}_{k}\right) p\left(\boldsymbol{z}_{k} \mid \boldsymbol{\mu}^{z}_{k}, \boldsymbol{\Sigma}^{z}_{k}\right) \mathrm{d} \boldsymbol{z}_{k}
\end{equation}\par
In general, (18) is not Gaussian since a Gaussian distributions mapped through a nonlinear function leads to a non-Gaussian predictive distribution \cite{deisenroth2010efficient}. Therefore, it can not be computed analytically. This issue is typically solved by approximation: Approximate (18) as a Gaussian distribution which has the same mean and variance function. Based on the criteria of computationally cheap and practical, a first order Taylor approximation method is adopted in this paper \cite{girard2003gaussian}.
\begin{equation}
\boldsymbol{\mu}_{k}^{d}=\boldsymbol{u}^{d}\left(\boldsymbol{\mu}_{k}^{z}\right), \boldsymbol{\Sigma}_{k}^{d}=\boldsymbol{\Sigma}^{d}\left(\boldsymbol{\mu}_{k}^{z}\right), \boldsymbol{\Sigma}_{k}^{d x}=\nabla_{x} \boldsymbol{\boldsymbol{\mu}}^{d}\left(\boldsymbol{\mu}_{k}^{z}\right) \boldsymbol{\Sigma}_{k}^{x}
\end{equation}

\section{GPMPC for Vehicle Obstacle Avoidance and Overtaking
Maneuvers}\label{sec3}
\subsection{Obstacle Avoidance and Overtaking Problems}

Obstacle avoidance is one of the most difficult maneuvers for an autonomous vehicle. It combines lateral and longitudinal motion of vehicles while avoiding collisions
with obstacles. In addition, other types of maneuvers such as lane-changing, lane-keeping and merging in a sequential manner should be considered \cite{dixit2018trajectory}. Overtaking can be treated as a moving-obstacle avoidance problem. The overtaking vehicle with faster speed is called ego vehicle while the vehicle to be overtaken with lower speed called lead vehicle. Fig. 2(a) and Fig. 2(b) illustrate the typical scenario of overtaking a static object and a dynamic object respectively.
\begin{figure}[H]
    \centering
    \includegraphics[scale=0.33]{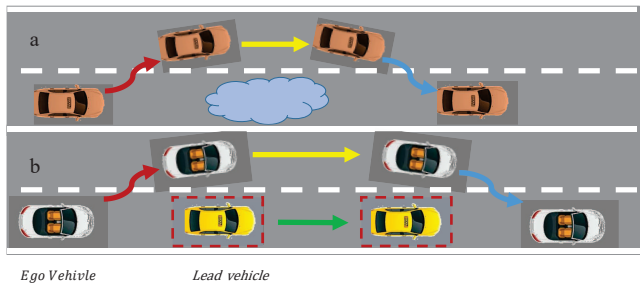}
     \caption{Typical scenario of overtaking
     (a) {Overtaking a static object}
     (b) {Overtaking a dynamic object}}
\end{figure}%

The essence of obstacle avoidance or overtaking problems are trajectory planning and trajectory tracking. The definitions of two terms have subtle differences that trajectory planning concerns more about how to generate a state trajectory, while tracking focuses on how to follow a planned trajectory. Basically, these two aspects are often studied together. In the literature there are a variety of approaches have been developed for collision avoidance and planning safe trajectories to overtake the obstacles. These methods can be grouped in four categories: graph-search based methods like rapidly exploring random trees \cite{kuwata2008motion}, artificial potential field based methods \cite{tang2010novel}, meta-heuristic based methods \cite{hussein2012metaheuristic} and mathematical optimization based
methods \cite{gao2010predictive}. Potential field based techniques are commonly used approaches since they have shown success in generating collision-free trajectories for overtaking \cite{kitazawa2016control}. However, they do not take the vehicle dynamics into account and hence can not ensure the reliability of the trajectories especially when the vehicle operates at high speed. \cite{dixit2018trajectory} proposed a method combines the potential field with MPC to overcome the absence of the vehicle model. This will turn the trajectory planning to several constraints that require to be satisfied.  A new problem arises because collision avoidance constraints are typically non-convex which will lead to the local minimum instead of global minimum of optimal problems. \cite{bengtsson2020autonomous} introduces learning model predictive control to approximate the issue but the new approach suffers from high computational complexity. \par
Another method proposed by \cite{franco2019short} is verified to be feasible to cope with obstacles avoidance. It also combines the adaptive MPC with collision avoidance methods. However, the bicycle model is
only concerned with kinematics, the lateral control with regard to tire model is simplified. The proposed method of this thesis is based on \cite{franco2019short} and extends to the more accurate vehicle models combined with data-driven control strategy, making it closer to the real scenarios. %
\subsection{Overtaking Scenario and Overtaking Constraints}

First and foremost, we need to build a scenario for overtaking problems. In this paper, we consider
the case of double lane change where overtaking maneuvers involved in it and it is also the most common cases in daily life, as shown in Fig. 2. The road width is set to 7.5m according to general highway standard. The solid black lines on both sides represent the road boundaries, the dashed line is the center line of the road. The ego vehicle drives from the left side to the right side and stays in the same lane all the time, unless there is an obstacle ahead that needs to overtake. There are a few lead vehicles or obstacles setting in front of the ego vehicle with constant velocity. On the contrary, the ego vehicle is given a greater degree of freedom and can adjust its speed in time according to the situation, such as accelerating when overtaking or braking when it needs to
maintain a safe distance from the lead vehicles.

The objective of overtaking problems is to maximize progress on the center line of the track and avoid collision at the same time, which is quite suitable for MPC controller. MPC controller can incorporate tracking constraints and overtaking constraints in a systematic way and make the controlled vehicle react in advance due to its long prediction horizon.

In order to avoid collision, there are some approaches relying on a higher-level path planner \cite{frazzoli2005maneuver,gray2012predictive}. However, this will lead to a rapid increase in computational complexity, which is not well suited for real-time overtaking. In our paper, we incorporate the path planning into the tracking controller by using additional constraints formed by the geometric relationship between the ego vehicle and other obstacle vehicles.

For safety overtaking maneuvers, we define an area called "safe zone" of the lead vehicle, which is a rectangle area around the lead vehicle. The safety zone is twice the length and width of the vehicle in length and width, respectively. The ego vehicle should not enter this area during the overtaking. At the next control interval the area is refreshed based on the new position of the lead vehicle. To avoid entering the area, the following state constraints are used:
$$\textbf{A}\boldsymbol{x} \leq \textbf{B}$$
where $\boldsymbol{x} = [X;Y;\Phi;v_x;v_y;\omega]$ is the state vector, while $\textbf{A}$ and $\textbf{B}$ are the constraint matrices that can be updated when the controller is in operation. Since the overtaking maneuvers are mainly related to the longitudinal and lateral motion, the constraints have effect on the position of the ego vehicle $(X,Y)$. The matrices $\textbf{A}$ and $\textbf{B}$ are defined as:
\begin{equation}\label{oc}
    \textbf{A}=\left[\begin{array}{cccccc}
0 & 1 & 0 & 0 & 0 & 0\\
0 & -1 & 0 & 0 & 0 & 0\\
k & -1 & 0 & 0 & 0 & 0\\
\end{array}\right], \quad \textbf{B}=\left[\begin{array}{c}
L_1 \\
L_2 \\
-b \\
\end{array}\right]
\end{equation}
where $k$ is the slope of the line formed from the c.g. of the ego vehicle to safe zone corner. Obviously, $b$ is the intercept. $L_{1,2}$ represent the upper bound and lower bound on the $Y$ coordinate. The coordinate system of the entire road is established with the origin point at the middle point of the left side of the road. Fig. 3 takes the left overtaking as an example, two vehicles are on the lower lane and the ego vehicle will detect the lead vehicle when the distance is less than $20 \textrm{m}$. The dashed orange area is the accessible area when overtaking happens, while the dashed red rectangle boundary is the safe zone of the lead vehicle. The following Algorithm 1 provides an overview of left overtaking algorithm.
$\epsilon$ is the extra safe lateral distance we added, which is set to half the width of the vehicle in this paper, as shown in Fig. 3.

\begin{algorithm}[ht]
  \caption{Left Overtaking}
  \label{alg1}
  \begin{algorithmic}[1]
   \Require
   The position of the ego vehicle;
   The position of the lead vehicle;
   \Ensure
   The constraint matrices: $\textbf{A}$ and $\textbf{B}$;
   \Function {LeftOvertakingConstraint}{Vehicle, detection, obstacle}
   \State $VehicleX = Vehicle(1), VehicleY = Vehicle(2)$;
   \State $LvRightLeftSafeX = obstacle(1)-obstaclelength$ ;
   \State $LvRightLeftSafeY = obstacle(2)+obstaclewidth$;
   \State $LvFrontLeftSafeX = obstacle(1)+obstaclelength$;
   \If {the lead vehicle is detected}
    \State $slope=(LvRightLeftSafeY-VehicleY)/(LvRightLeftSafeX-VehicleX)$
    \If {$VehicleX \leq LvRightLeftSafeX$}
        \If{$VehicleY > LvRightLeftSafeY$}
            \State $k=0$; $b=LvRightLeftSafeY+\epsilon$;
        \Else
            \State $k = slope$;
            \State $b = LvRightLeftSafeY-k*LvRightLeftSafeX$
        \EndIf
    \If {the ego vehicle is parallel to the lead vehicle}
        \State $k=0$; $b=LvRightLeftSafeY+\epsilon$;
    \Else
        \State $k=0$; $b=-L_2$;

    \EndIf
    \EndIf
   \Else
    \State  $k=0$; $b=-L_2$;
   \EndIf
   \State
    \Return{$\textbf{A},\textbf{B}$}
   \EndFunction
  \end{algorithmic}
\end{algorithm}

\textit{Remark:} When the ego vehicle needs to overtake, the choice of left overtaking or right overtaking can be determined simply according to the position of the front vehicle in the coordinate system of the ego vehicle. If the Y coordinate of the front vehicle is negative, then the ego vehicle will choose left overtaking, otherwise choose right overtaking.
\begin{figure}[ht]
    \centering
    \includegraphics[scale=0.4]{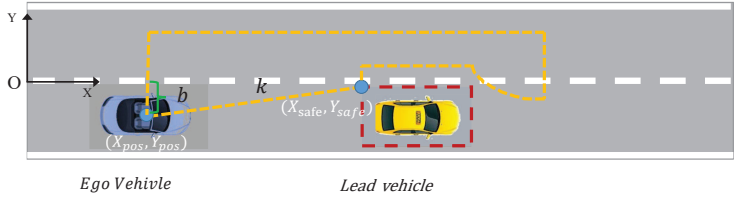}
     \caption{Schematic of accessible driving area in the case of left overtaking}\label{cono}
\end{figure}%
\subsection{Contouring Control and Resulting Cost Function}

The GP-based MPC controller makes use of a contouring control framework, which follows a similar strategy as used in \cite{lam2010model}. We adapt the particular formulation of the model predictive contouring control (MPCC) to maximize the travelled distance on the reference path. Therefore, the center line of a certain lane is chosen as the reference path, but it is employed merely as a measure of progress. The reference path is parameterized by its arc length $\xi$ using third order spline polynomials. Then given an exemplary $\xi$, the centerline position $[X_{c}(\xi),Y_{c}(\xi)]$ and orientation $\Phi_{c}(\xi)$ and the track radius $R_{c}(\xi)$ of vehicle can be evaluated by interpolation. As a result, the cost function is defined by the so-called lag error $e_l$, contour error $e_c$, orientation error $e_o$ and offset error $e_{off}$, as illustrated in Fig. 4 and defined as follows:
\begin{equation}
    \begin{aligned}
e_{l}\left(\boldsymbol{u}^{x}_{k}, \xi_{k}\right)=&\textrm{cos} \left(\Phi\left(\xi_{k}\right)\right)\left(X_{c}\left(\xi_{k}\right)-X_{k}\right) \\
&+\textrm{sin} \left(\Phi\left(\xi_{k}\right)\right)\left(Y_{c}\left(\xi_{k}\right)-Y_{k}\right) \\
e_{c}\left(\boldsymbol{u}^{x}_{k}, \xi_{k}\right)=& -\textrm{sin} \left(\Phi\left(\xi_{k}\right)\right)\left(X_{c}\left(\xi_{k}\right)-X_{k}\right) \\
&+\textrm{cos} \left(\Phi\left(\xi_{k}\right)\right)\left(Y_{c}\left(\xi_{k}\right)-Y_{k}\right)\\
e_o\left(\boldsymbol{u}^{x}_{k}, \xi_{k}\right)=&1-|\textrm{cos}\left(\Phi\left(\xi_{k}\right)\right)\cos\left(\varphi\right)+ \textrm{sin}\left(\Phi\left(\xi_{k}\right)\right)\textrm{sin}\left(\varphi\right)|\\
e_{off}\left(\boldsymbol{u}^{x}_{k}, \xi_{k}\right)=& \frac{1}{R_{c}\left(\xi_{k}\right)}\sqrt{e_{l}\left(\boldsymbol{u}^{x}_{k}, \xi_{k}\right)^2 + e_{c}\left(\boldsymbol{u}^{x}_{k}, \xi_{k}\right)^2}-1
\end{aligned}
\end{equation}

The MPC formulation can be made more efficient by removing constraints. However, to keep the vehicle staying inside the boundary of road, there must be vehicle state constraint. In this paper, we transform the traditional hard state constraint into a soft constraint and incorporate it as relaxed barrier function $\mathcal{R}_b(e_{off})$ into the cost function, which will improve the optimizer performance. The relaxed barrier function is defined as:
\begin{equation}
    \mathcal{R}_b(e_{off})=\beta\left(\sqrt{\frac{\left(c+\gamma(\lambda-e_{off})^{2}\right)}{\gamma}}-(\lambda-\boldsymbol{x})\right)
\end{equation}
where $\beta$, $\gamma$, $\lambda$ and $c$ are constant parameters. The stage cost function is then written as:
\begin{equation}\label{eq5.6}
\begin{aligned}
    l\left(\boldsymbol{u}^{x}_{k},\xi_{k}\right)=& \left\|e_{c}\left(\boldsymbol{u}^{x}_{k}, \xi_{k}\right)\right\|_{q_c}^{2} + \left\|e_{l}\left(\boldsymbol{u}^{x}_{k}, \xi_{k}\right)\right\|_{q_l}^{2} \\
    &+ \left\|e_{o}\left(\boldsymbol{u}^{x}_{k}, \xi_{k}\right)\right\|_{q_o}^{2} + \left\|\mathcal{R}_b(e_{off}\left(\boldsymbol{u}^{x}_{k}, \xi_{k})\right)\right\|_{q_{off}}^{2}
    \end{aligned}
\end{equation}
where $q_c$, $q_l$, $q_o$ and $q_{off}$ are the corresponding weights.

\begin{figure}[H]
    \centering
    \includegraphics[scale=0.35]{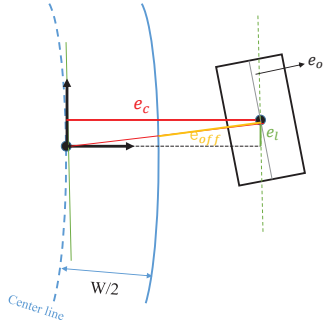}
     \caption{Lag-, contour-, orientation- and offset error. The vehicle model is intentionally plotted outside the road boundary to show these errors clearly. }
\end{figure}%

\subsection{Input Constraints and Resulting Formulation}
The input vector constraints $\mathcal{U}$ are limited as below:
\begin{equation}
    \left[\begin{array}{c}
-\delta_{\max } \\
-T_{\max } \\
\end{array}\right] \leq\left[\begin{array}{c}
\delta_{k} \\
T_{k} \\
\end{array}\right] \leq\left[\begin{array}{c}
\delta_{\max } \\
T_{\max } \\
\end{array}\right]
\end{equation}

Based on this contouring formulation, we integrate a stochastic GP-based MPC model which results in minimizing the cost function (23) over a finite horizon of length $N_p$. The corresponding GP-based MPC formulation with tractable approximation is defined as follows:
\begin{subequations}\label{GPMPC}

\begin{align}
    \min _{ \boldsymbol{u}_{k}} \quad & J \left(\boldsymbol{\mu}_{k}^{x},\xi_{k}\right) =\sum_{k=0}^{N-1} l\left(\boldsymbol{\mu}_{k}^{x}, \xi_{k}\right) \label{GPMPCa}\\
     s.t. \quad
     &  \boldsymbol{u}^{x}_{k+1}=\boldsymbol{f}_{n}\left(\boldsymbol{u}^{x}_{k}, \boldsymbol{u}_{k}\right)+\textbf{B}_{d} (\boldsymbol{d}\left(\boldsymbol{u}^{x}_{k}, \boldsymbol{u}_{k}\right)+\boldsymbol{w}_k) \label{GPMPCb}\\
     & \textbf{A} \boldsymbol{\mu}_{k+1}^{x} \leq \textbf{B} \label{GPMPCc}\\
     & \boldsymbol{u}_{k} \in \mathcal{U} \label{GPMPCd}\\
     & \boldsymbol{\mu}_{0}^{x} = \boldsymbol{x}(k), \boldsymbol{\Sigma}_{0}^{x} =0, \xi_{0}=\xi(k) \label{GPMPCe}
\end{align}
\end{subequations}\par

\section{Simulation and Analysis}

In order to verify the effectiveness of the proposed approach, a left overtaking scenario on a two-lane road  is constructed. There are two vehicles traveling at different speeds in front of the ego vehicle with same travelling direction in the same lane. The speed of the ego vehicle is faster than the other two vehicles. The nonlinear MPC (NMPC) algorithm is used for comparison. The GPMPC problem in (25) is implemented with a prediction horizon of $N_p=10$. The sampling time is $T_s=50\,\textrm{ms}$, resulting in a $0.5\,\textrm{s}$ look-ahead. The maximum number of iterations is limited to 30 to ensure consistent maximum solve times. Considering the reality, we limit the vehicle speed between $10\,\textrm{m/s}$ to $35\,\textrm{m/s}$. The vehicles and obstacles are abstracted to small blocks with $4\,\textrm{m}$ long and $1.6\,\textrm{m}$ wide. For easy distinction, vehicles are colored. The ego vehicle is black, the obstacle and lead vehicles are depicted in red or green. The parameters of the ego vehicle is shown in Table 1. For the full MAGIC formulas, $B_f$ is 0.4, $C_f$ is 8, $D_f$ is 4560.4, $E_f$ is -0.5, $B_r$ is 0.45, $C_r$ is 8, $D_r$ is 4000, $E_r$ is -0.5.

\begin{table}[H]
\caption{Parameters of the ego vehicle}
\label{tab:1}       
\begin{tabular}{llllll}
\hline\noalign{\smallskip}
$M (kg)$ & $I_z (kg \cdot m^2)$ & $L_f (m)$ & $L_r (m)$ & $C_{l,f} (N/rad)$& $C_{l,r} (N/rad)$ \\
\noalign{\smallskip}\hline\noalign{\smallskip}
500 & 600 & 0.9 & 1.5 & 1400 & 1400\\
\noalign{\smallskip}\hline
\end{tabular}
\end{table}
We compare two MPC-based controller, NMPC and GPMPC. Nominal vehicle model $\boldsymbol{f}_{n}\left(\boldsymbol{x}_{k}, \boldsymbol{u}_{k}\right)$ and true vehicle model (12) are prepared for calculating the deviations to be learned by GP model. At first, the ego vehicle starts at the initial point with NMPC controller, meaning that all GP-dependent variables are set to zero. The corresponding parameters are tuned slightly to prevent crashes and driving off the road. Since the nominal model does not consider the model mismatch and unmodeled dynamics, it is allowed that its driving behaviour is somewhat erratic and there are small collisions with road boundaries. Throughout the first step, the data are collected to fill the trainning dictionary and train the GP error model $\boldsymbol{d}$. The GP model was first generated with fixed hyperparameters, but we can infer the hyperparameters by using maximum likelihood optimization according these collected data \cite{rasmussen2003gaussian}. After that we activate the GPMPC with loaded data and optimized hyperparameters. The new training data will be added into the GP model after each iteration. When the maximal dictionary size is reached, some data points will be discarded by using method mentioned in (14).

\begin{table}[H]
\caption{Parameters of the NPMC controller}
\label{tab:2}       
\begin{tabular}{lllllllll}
\hline\noalign{\smallskip}
$q_c$ & $q_l$ & $q_o$ & $q_{off}$ & $\beta$ & $c$& $\gamma$ & $\lambda$ &$u_{max}$ \\
\noalign{\smallskip}\hline\noalign{\smallskip}
20 & 50 & 20 & 180 & 1000& 5& 4&  -0.1 & $[0.3419 \quad 1]^\textrm{T}$\\
\noalign{\smallskip}\hline
\end{tabular}
\end{table}

The initial position of the ego vehicle is set to $(0,-1.875)$ with an initial speed $20\,\textrm{m/s}$. Lead vehicle 1 starts from $(25,-1.875)$ with constant velocity $12\,\textrm{m/s}$, while lead vehicle 2 is at point $(60,-1.875)$ with constant velocity $10\,\textrm{m/s}$. The global coordinate system is shown in Fig. 3. The parameters of the NPMC controller is shown in Table 2. For GPMPC controller, its parameters in the MPC part are the same as those of NPMC. The hyper-parameters of GPMPC controller are shown in Table 3, where
$\textbf{M}_1$, $\sigma_{v_{x}}^{2}=7.1304e-4$ and $\sigma_{f1}=2.8052e-11$ are the hyper-parameters for $v_x$ dimension, $\textbf{M}_2$, $\sigma_{v_{y}}^{2}=1.0358e-10$, and $\sigma_{f2}=0.0236$ are the hyper-parameters for $v_y$ dimension, $\textbf{M}_3$, $\sigma_{\omega}^{2}=1.0059e-10$ and $\sigma_{f3}=0.0117$ are the hyper-parameters for $\omega$ dimension.

\begin{table}[H]
\caption{Hyper-parameters for GPMPC controller}
\label{tab:3}       
\begin{tabular}{lll}
\hline\noalign{\smallskip}
Parameter  &Value  \\
\noalign{\smallskip}\hline\noalign{\smallskip}
$\textbf{M}_1$  &diag(0.0346,0.0151,0.0148,0.0153,0.0163,0.0156,0.0148,0.016) \\
$\textbf{M}_2$  &diag(9.9184e4,6.94995e4,731,1988,15,6.2355e4,0.12,1098)  \\
$\textbf{M}_3$  &diag(9.9829e4,9.6999e4,1.199e4,2131,77,12,0.51,982)   \\
\noalign{\smallskip}\hline
\end{tabular}
\end{table}

To quantify the performance of the GPMPC control scheme and the improvement due to the learning, we compare the predicted model error in $v_x$, $v_y$ and $\omega$, calculated by nominal model with NMPC controller and estimated model with GPMPC controller respectively.

Fig. 5 illustrates that GPMPC performs much better than NMPC. In order to see the capability of GP learning model clearly, the mean squared error (MSE) of the tracking error in each dynamic state are shown in Table 4.
\begin{figure}[H]
    \centering
    \includegraphics[scale=0.295]{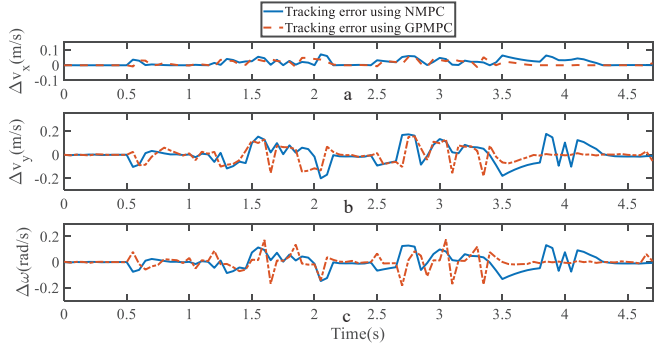}
     \caption{Tracking error using NMPC and GPMPC for left overtaking
     (a) {Tracking error $v_x$}
     (b) {Tracking error $v_y$}
     (c) {Tracking error $\omega$}}
\end{figure}%

\begin{table}[H]
\caption{Comparison of tracking performance between NMPC and GPMPC}
\label{tab:4}       
\begin{tabular}{lllll}
\hline\noalign{\smallskip}
Model & $\overline{\left\|\mathbf{e}_{v_x}\right\|}$ & $\overline{\left\|\mathbf{e}_{v_y}\right\|}$ & $\overline{\left\|\mathbf{e}_{\omega}\right\|}$ & $\overline{\left\|\mathbf{e}\right\|}$ \\
\noalign{\smallskip}\hline\noalign{\smallskip}
NMPC & 0.2700 & 0.7684 & 0.5693 & 0.9565\\
GPMPC & 0.2025 & 0.6494 &  0.5659 & 0.8000\\
\noalign{\smallskip}\hline
\end{tabular}
\end{table}
Where $\left\|\boldsymbol{e}_{NMPC}\right\|=\left\|\boldsymbol{x}_{k+1}-\boldsymbol{f}\left(\boldsymbol{x}_{k}, \boldsymbol{u}_{k}\right)\right\|$,
$\left\|\boldsymbol{e}_{GPMPC}\right\|=\left\|\boldsymbol{x}_{k+1}-\left(\boldsymbol{f}\left(\boldsymbol{x}_{k}, \boldsymbol{u}_{k}\right)+\mathbf{B}_{d} \boldsymbol{\mu}^{d}\left(\boldsymbol{z}_{k}\right)\right)\right\|$

In addition, we investigate the controller performance by plotting the predicted values in one iteration. In each time step, both controllers will make predictions for 10 steps ahead as shown in Fig. 6-Fig. 11.

\begin{figure}[H]
    \centering
    \includegraphics[scale=0.295]{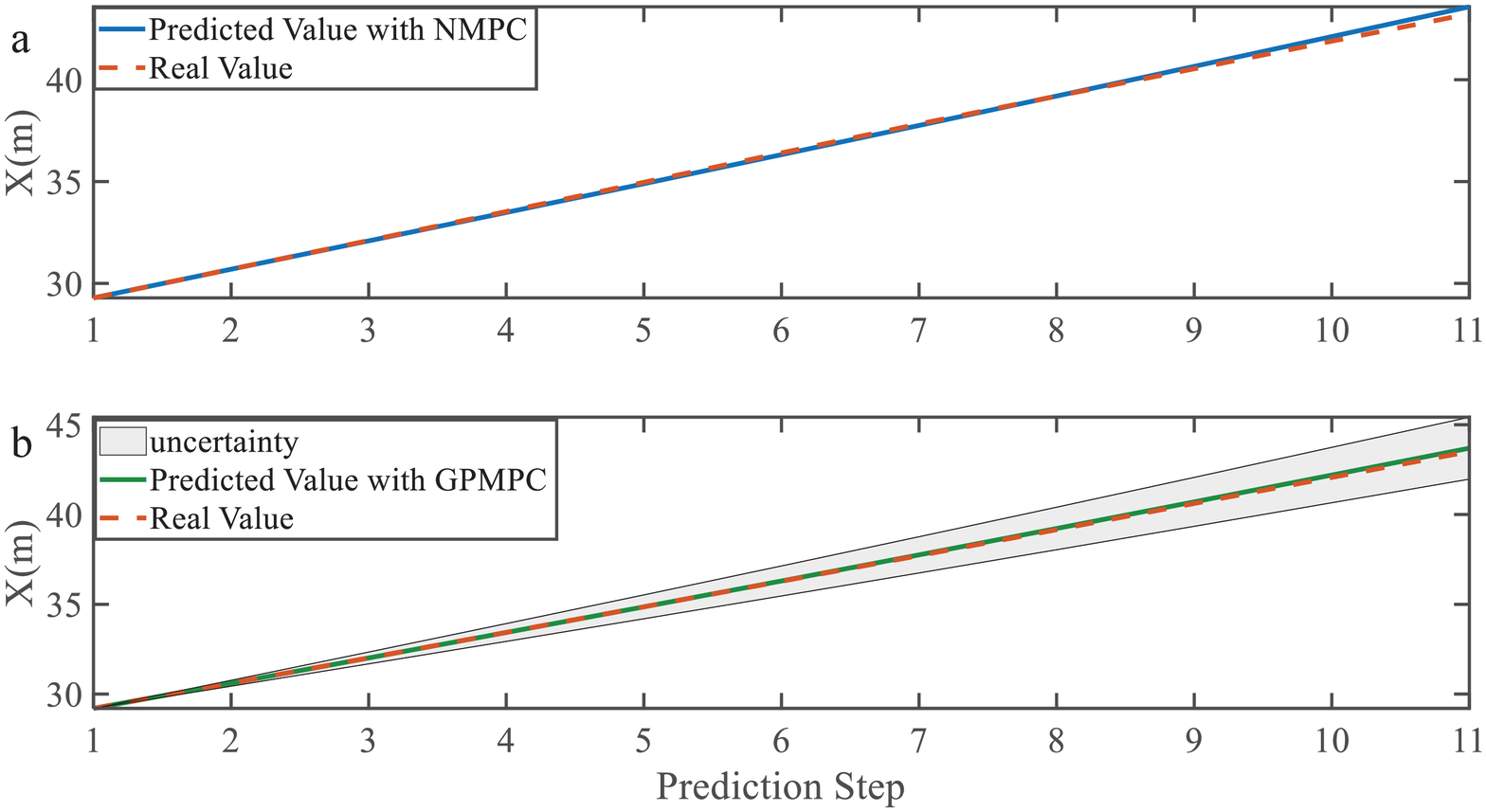}
     \caption{Position X evolution for multi-step prediction
     (a) {NMPC}
     (b) {GPMPC}}
\end{figure}%

\begin{figure}[H]
    \centering
    \includegraphics[scale=0.295]{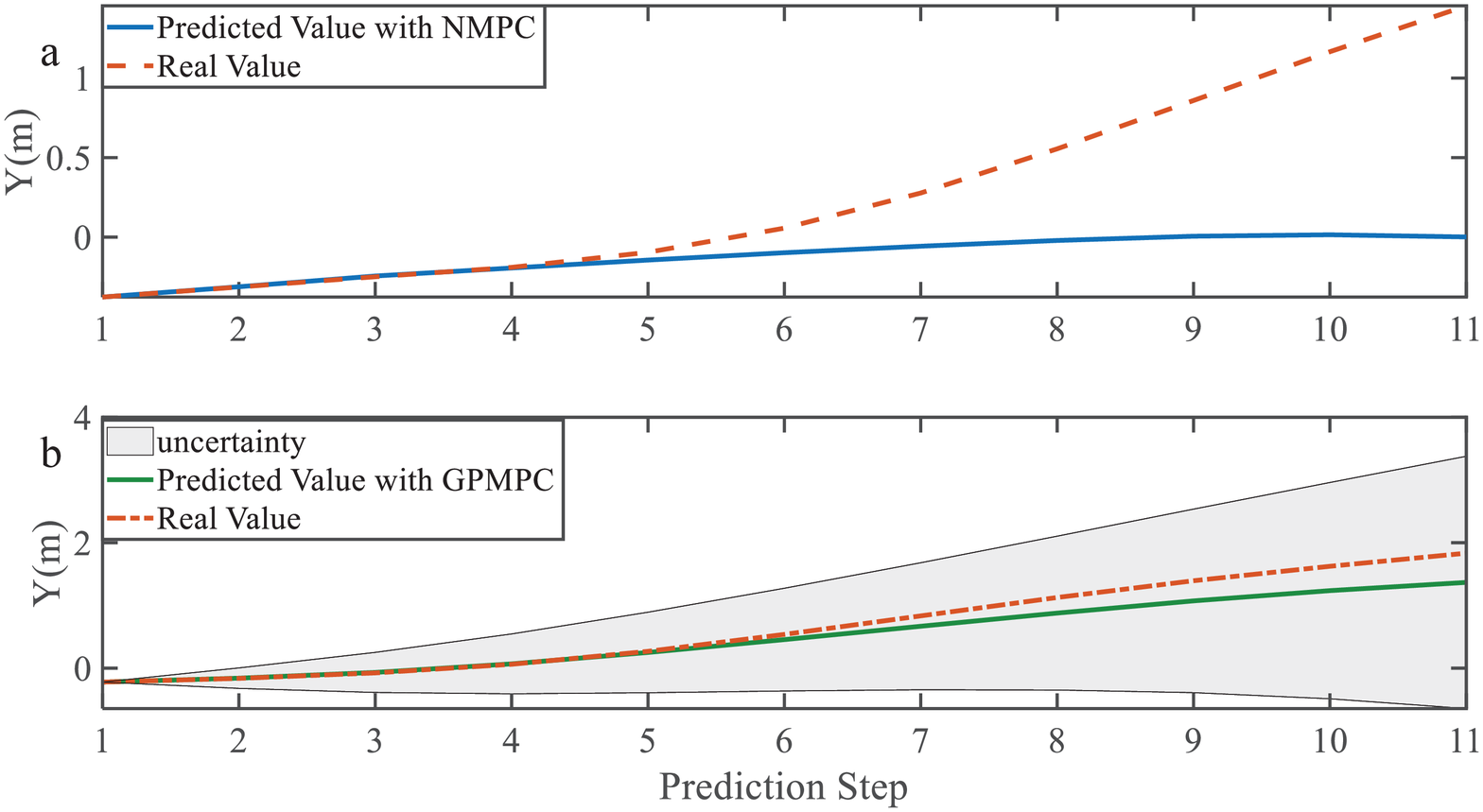}
     \caption{Position Y evolution for multi-step prediction
     (a) {NMPC}
     (b) {GPMPC}}
\end{figure}%

As evident in Fig. 6-Fig. 11, the NMPC controller performs visually suboptimally and is unable to predict the future evolution in some cases. On the contrary, we can see that the GPMPC controller matches the real values quite well in most cases. The uncertainty in form of a $2\sigma$ confidence interval is shown in light gray. With uncertainty propagation, we observe that the majority of predictive states during overtaking are still anticipated by the GP-uncertainty.

\begin{figure}[H]
    \centering
    \includegraphics[scale=0.295]{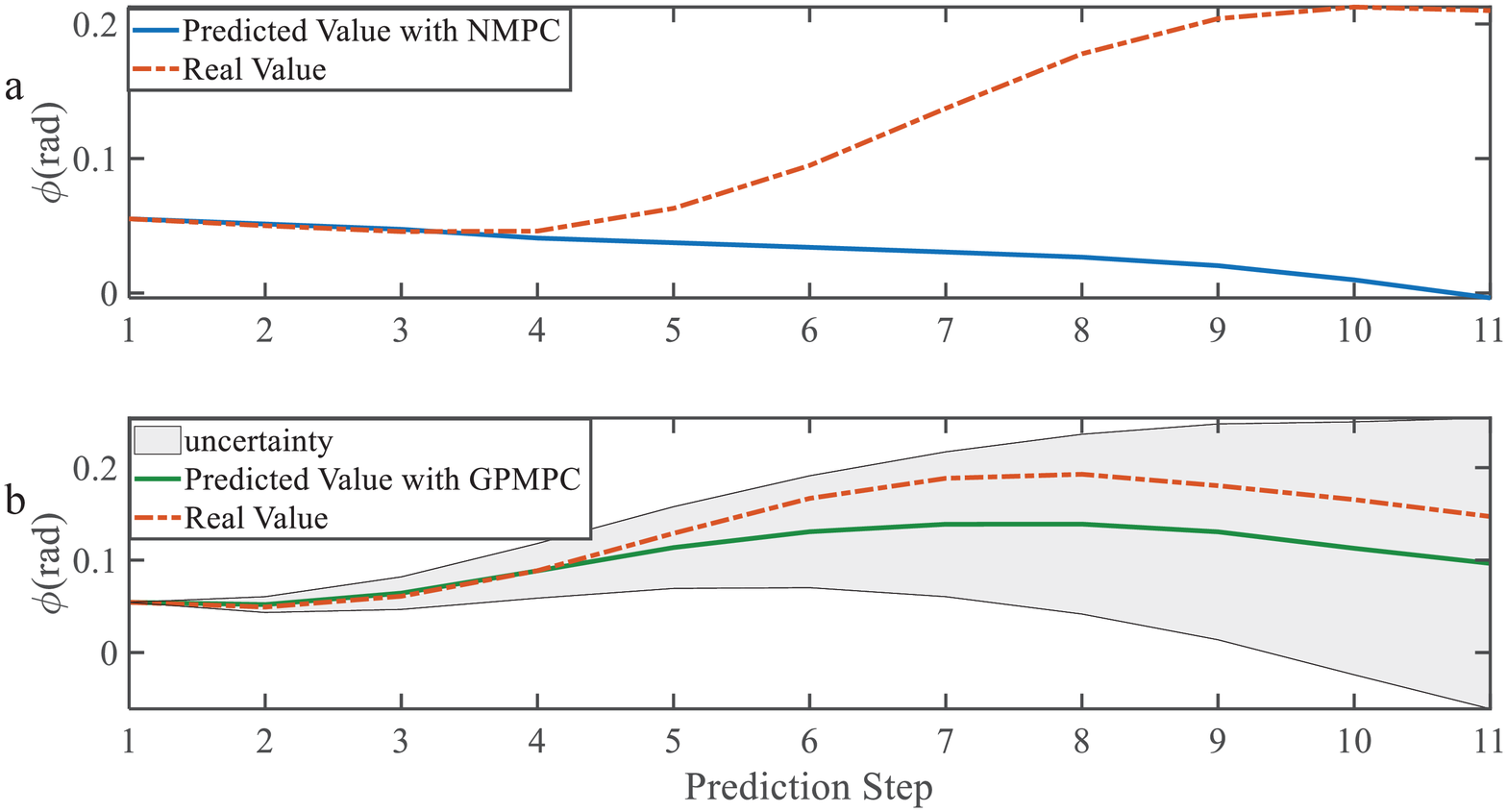}
     \caption{Yaw angle evolution for multi-step prediction
     (a) {NMPC}
     (b) {GPMPC}}
\end{figure}%
\begin{figure}[H]
    \centering
    \includegraphics[scale=0.295]{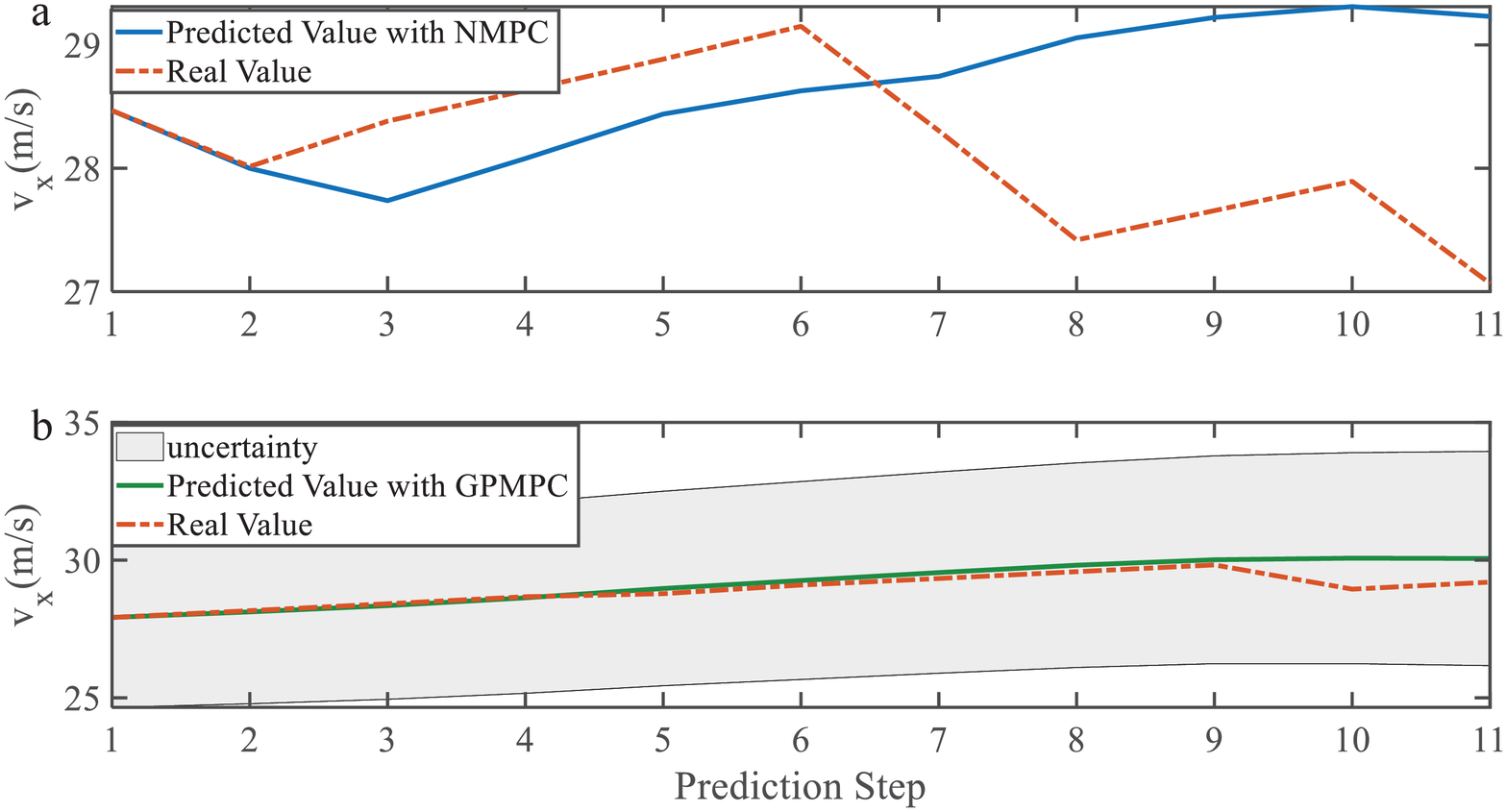}
     \caption{Velocity $\boldsymbol{v}_x$ evolution for multi-step prediction
     (a) {NMPC}
     (b) {GPMPC}}
\end{figure}%

\begin{figure}[H]
    \centering
    \includegraphics[scale=0.295]{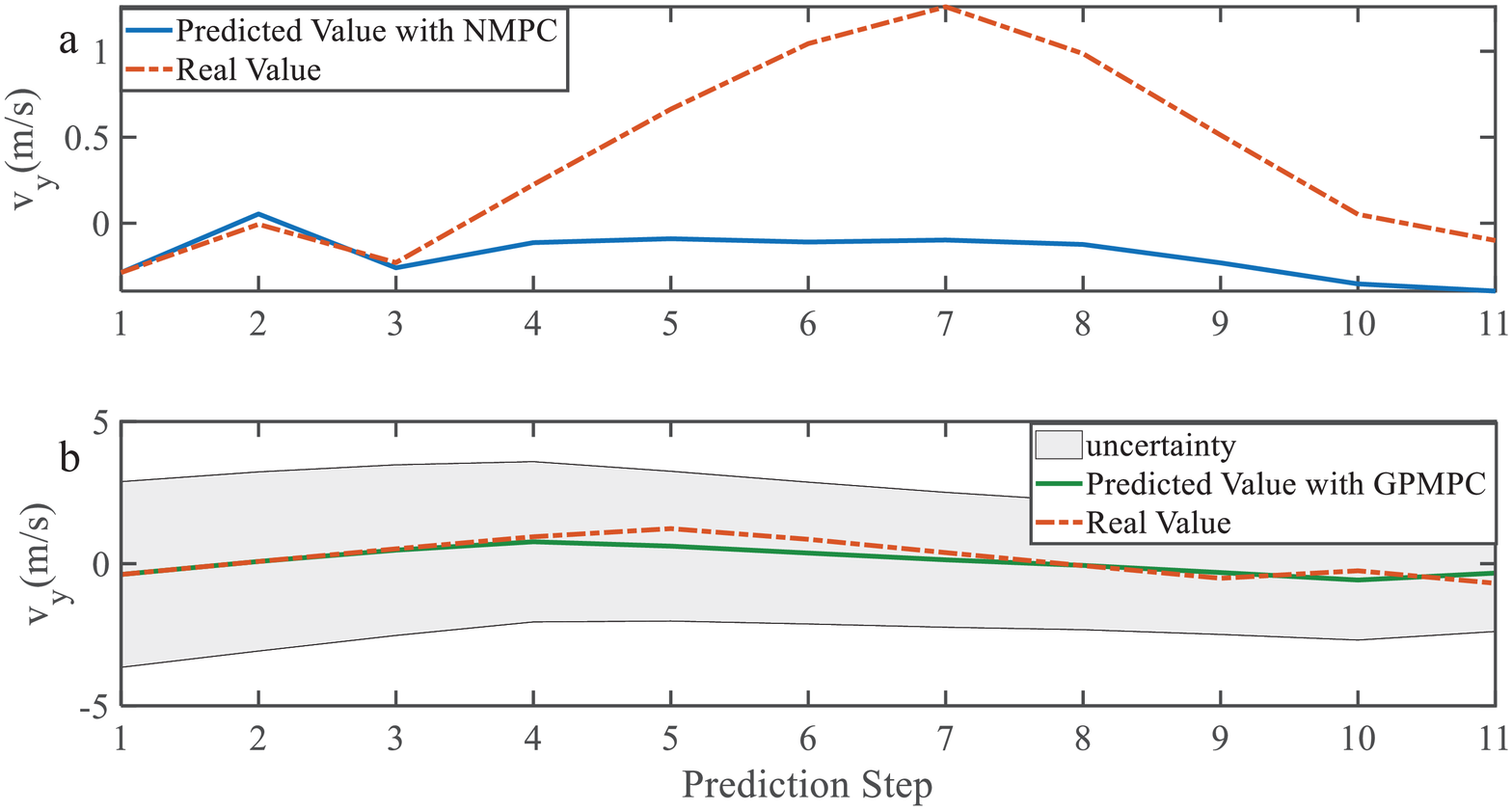}
     \caption{Velocity $\boldsymbol{v}_y$ evolution for multi-step prediction
     (a) {NMPC}
     (b) {GPMPC}}
\end{figure}%
\begin{figure}[H]
    \centering
    \includegraphics[scale=0.295]{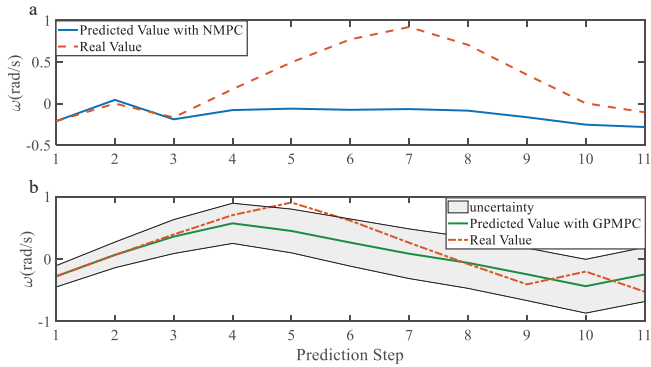}
     \caption{Yaw rate evolution for multi-step prediction
     (a) {NMPC}
     (b) {GPMPC}}
\end{figure}%

Evolution of control inputs throughout the whole simulation is another important index for controller performances. Fig. 12 shows applied control inputs to left overtaking problem. The upper figures represent the change of the first control variable: steering angle $\delta$. The green dashed line is GPMPC. During the time interval $1.5-2\,\textrm{s}$ and $3-3.5\,\textrm{s}$ in Fig. 12(a), $\delta$ changed more rapidly than in NMPC, which means GPMPC consumed a lot to achieve a steady state.

When it comes to the control variable $T$, GPMPC performs much better than NMPC. Since $T=1$ represents full accelerating and $T=-1$ means full braking, Fig. 12(b) shows that NMPC controller shifts extremely steep to avoid constraint violation. This limitation is not present in the GPMPC approach, where a more precisely prediction has made. Therefore, except a few points where the vehicle first detect the obstacle, GPMPC controls fairly smooth comparing to the NMPC controller, which gives a speed benefit and consumes less power.

\begin{figure}[H]
    \centering
    \includegraphics[scale=0.295]{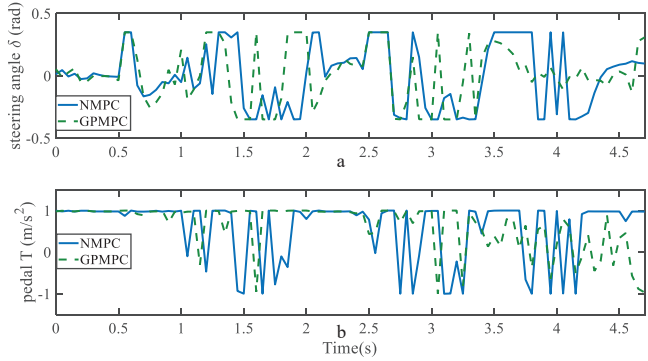}
     \caption{Control inputs using NMPC and GPMPC for left overtaking
     (a) {Steering angle}
     (b) {Pedal}}
\end{figure}%

For autonomous overtaking scenario, taking driven trajectories as a performance criterion is the most intuitive way. Therefore, we investigate two controller performances by comparing the overtaking maneuvers and the overall driven trajectories. Fig. 13(a) and Fig. 13(c) are the driven trajectories with velocity profile generated by NMPC and GPMPC respectively. The maneuvers where the the ego vehicle is overtaking the first vehicle are shown in Fig. 13(b) and Fig. 13(d). Both control strategies are able to accomplish the overtaking mission without collisions, which proves that the parameters of MPC controller are valid. However, it is evident to see GPMPC outperforms NMPC, especially with regard to constraint satisfaction. This can be seen from the black dotted circle in Fig. 13(b). Although the "safe zone", depicted by dashed red lines, is virtual in real world, driving too close to the overtaken vehicles  will indeed increase the risk of collision.

Furthermore, there is another phenomenon worth mentioning. It can be seen from NMPC's trajectories that there is a wave crest after overtaking the second vehicle, which is clearly visible from the blue dotted circle in Fig. 13(a). The extra displacement of trajectories are generated due to the wrongly estimation of lateral force by NMPC, which leads the vehicle return to the original track prematurely when it not fully finishes overtaking. Then, since the lead vehicle is moving, at the next time step, situation gets not suitable for overtaking. The ego vehicle has to overtake the obstacle vehicle again.
Comparing to NMPC, the resulting traces with GPMPC are displayed in Fig. 13(c), generally showing a much more smooth and safe overtaking behavior. In particular, almost all of the problems in the trajectories of the nominal model and NMPC controller can be alleviated.
\begin{figure}[H]
    \centering
    \includegraphics[scale=0.35]{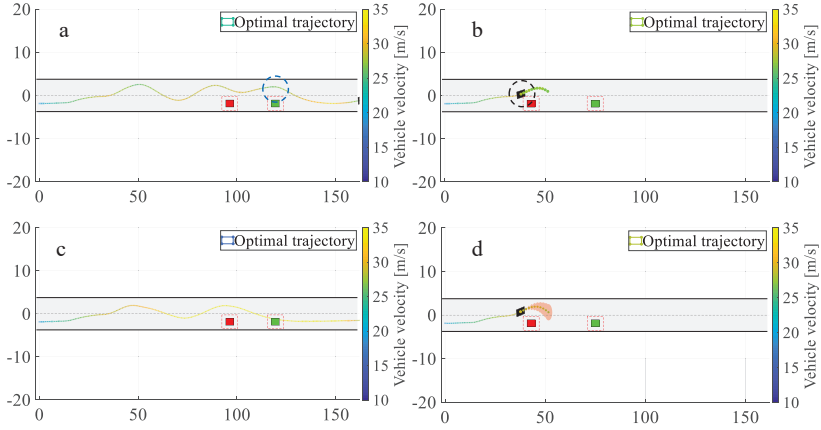}
     \caption{Overtaking maneuvers and the overall driven trajectories
     (a) {Overall driven trajectories of NMPC}
     (b) {Overtaking maneuver of NMPC}
     (c) {Overall driven trajectories of GPMPC}
     (d) {Overtaking maneuver of GPMPC}}
\end{figure}%

\section{Conclusion}\label{sec10}

We have investigated overtaking problems in autonomous driving and dedicate to build a GP-based control framework which is able to complete vehicle control, trajectory tracking and obstacle avoidance.  Since the vehicle model is a extremely complicated system and the road condition is time-varying, it is intractable to derive a precise model. Thus, the learning based method is introduced and the core concept of this method is only using a nominal model to represent the vehicle while the rest uncertainties, disturbances and mismatch can be learned by GP model. However, one issue raised during the combination of GP regression and traditional NMPC controller: the MPC became a stochastic formulation because of the GP approximation. By employing the Taylor approximation we can propagate the uncertainties and evaluate the residual uncertainties, which increases the accurateness and the controller. The implemented Taylor approximation depends directly on the dimension of training data. As the data points constantly adding into the model, it becomes expensive to evaluate in high dimensional spaces. We limit the upper bound of the number of data points with a dictionary and set a selection mechanism, thus the computational complexity will be sustained on a medium level. In addition, we modify the constraints and cost function to reduce the computation need for optimization. Collectively, simulation results demonstrate that both performance and safety in overtaking can be improved by using GPMPC.

\section{Declarations}
\subsection{Funding}
This research has received funding from the European Unions Horizon 2020 Framework Programme for Research and Innovation under the Specific Grant Agreement No. 945539 (Human Brain Project SGA3), from the National Natural Science Foundation of China (No. 61906138), and from the Shanghai AI Innovative Development Project 2018.
\bibliographystyle{unsrt}
\bibliography{test}

\end{document}